\documentclass{article} 
\usepackage{acra}
\usepackage{todonotes}

\usepackage[utf8]{inputenc} 
\usepackage[T1]{fontenc}    
\usepackage{url}            
\usepackage{booktabs}       
\usepackage{amsfonts}       
\usepackage{nicefrac}       
\usepackage{microtype}      
\usepackage{graphicx}
\usepackage{subcaption}
\usepackage{amsmath}
\usepackage{amssymb}
\usepackage{pifont}
\usepackage{textcomp}
\usepackage{gensymb}
\usepackage{bm}
\usepackage{xcolor}
\usepackage{tikz}
\usepackage{hyperref}
\usepackage{algorithm,algorithmic}

\title{\LARGE \bf
Evaluating task-agnostic exploration for fixed-batch learning of \\arbitrary future tasks
}

\author{Vibhavari Dasagi$^{1}$, Robert Lee$^{1}$, Jake Bruce$^{1,2}$, and J\"urgen Leitner$^{1}$ \\ $^{1}$Queensland University of Technology (QUT), Brisbane, Australia
\\ $^{2}$DeepMind, London, UK
\\ Contact: vibhavari.dasagi@hdr.qut.edu.au\\}

\begin{document}

\maketitle
\thispagestyle{empty}
\pagestyle{empty}

\begin{abstract}

Deep reinforcement learning has been shown to solve challenging tasks where large amounts of training experience is available,
usually obtained online while learning the task.
Robotics is a significant potential application domain for many of these algorithms, but generating robot experience in the
real world is expensive, especially when each task requires a lengthy online training procedure.
Off-policy algorithms can in principle learn arbitrary tasks from a diverse enough fixed dataset.
In this work, we evaluate popular exploration methods by generating robotics datasets for the purpose of learning
to solve tasks completely offline without any further interaction in the real world. We present results on three popular continuous control tasks in simulation, as well as
continuous control of a high-dimensional real robot arm. Code documenting all algorithms, experiments, and hyper-parameters is available
at \url{https://github.com/qutrobotlearning/batchlearning}.

\end{abstract}

\section{Introduction}

Recent research in the field of model-free deep reinforcement learning (RL) has enabled complex, expressive policies to be learned
from experience for many challenging simulated and virtual task domains~\cite{mnih2015human,lillicrap2015continuous}.
The success of these methods suggests potential applications to robotics, and some progress has been made in this
direction~\cite{frank2014curiosity,zhang2015towards,rusu2016sim,levine2016end,gu2017deep,vevcerik2017leveraging}.
A key limitation in learning robotics skills with deep reinforcement learning is the cost of gathering new experience. 
Since different control tasks with the same robot often involve similar observations, actions, and dynamics, it would
be convenient to gather a single dataset with diversity sufficient for agents to learn to solve arbitrary
future tasks completely offline; this setting is known as
\textit{batch reinforcement learning}~\cite{watkins1992q,ernst2005tree,lange2012batch,liu2015feature}: a well known
classical paradigm that has been relatively under-explored in the context of modern robotics research.

\begin{figure}
	\centering
	\begin{subfigure}[b]{0.45\linewidth}
	    \includegraphics[width=\linewidth]{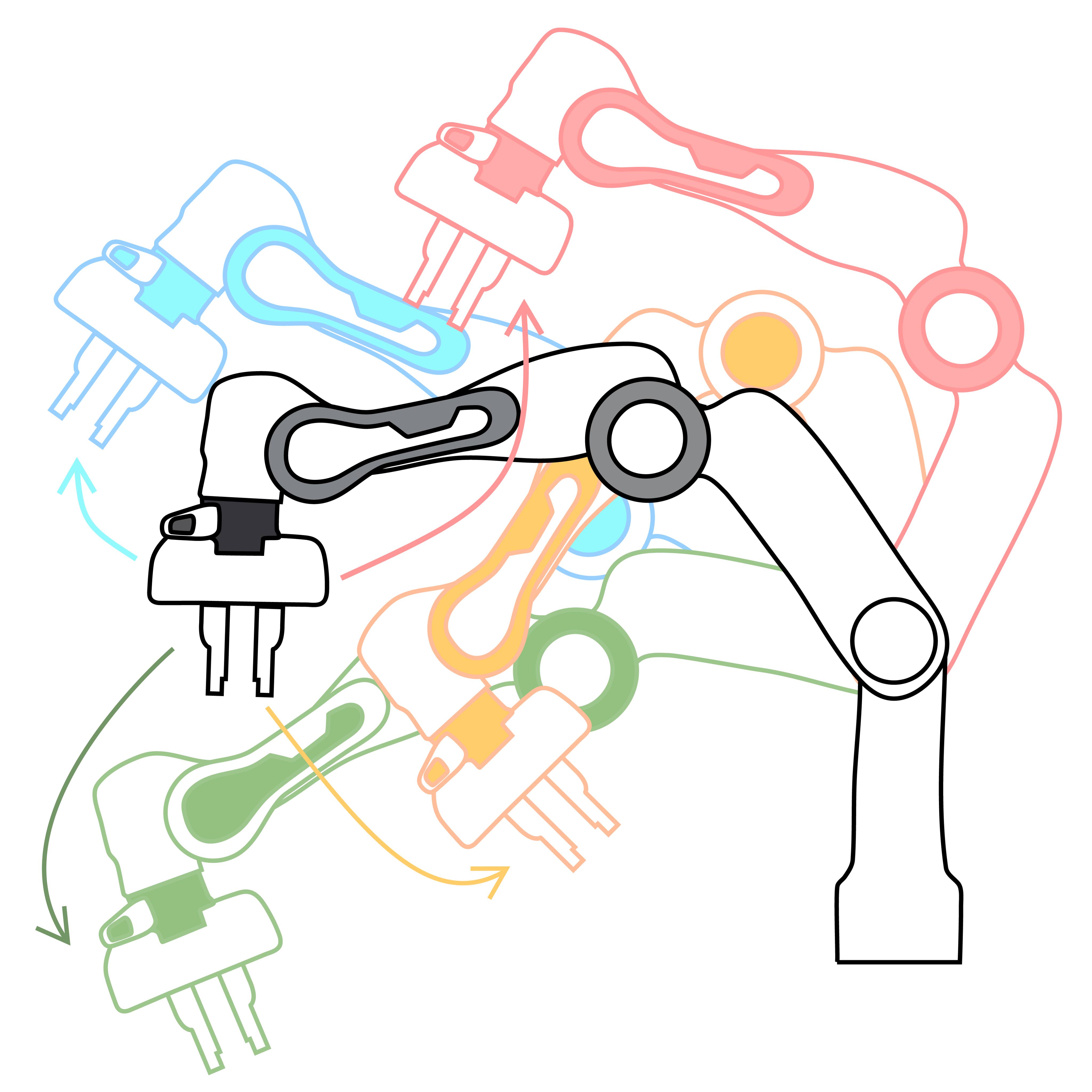}
	    \subcaption{Exploration phase}
	\end{subfigure}
	\begin{subfigure}[b]{0.45\linewidth}
    	\includegraphics[width=\linewidth]{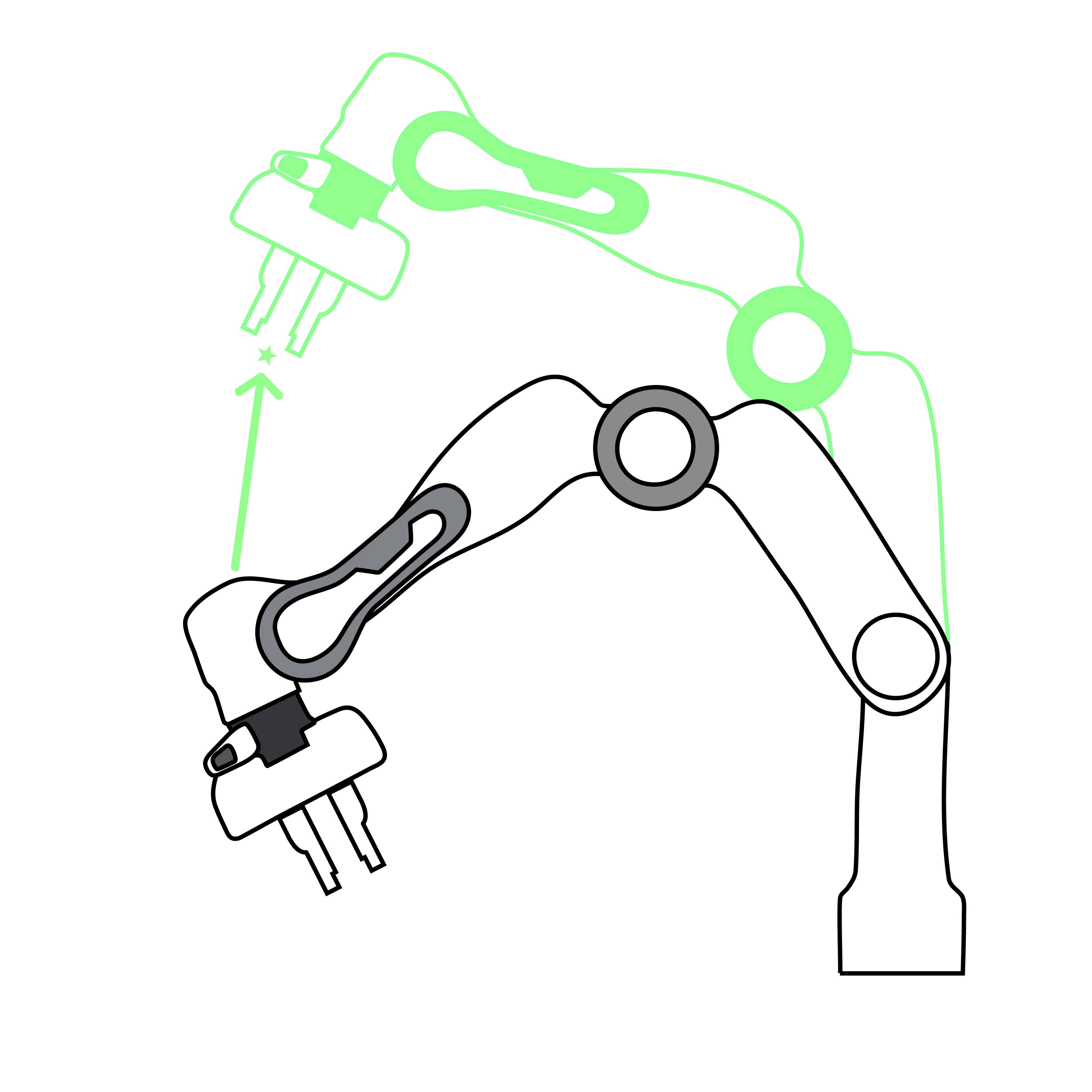}
    	\subcaption{Learning phase}
	\end{subfigure}
	\caption{In this work, we separate the phases of data gathering and policy learning.
	We evaluate the performance of state-of-the-art exploration
	methods by using the data they collect to learn to solve arbitrary tasks completely offline.}
    \label{fig:abstract}
    \vspace{-0.3cm}
\end{figure}

While many contemporary approaches can be described as \emph{on-policy} in which all the training data in each update
is generated directly by the current version of the agent being trained~\cite{mnih2016asynchronous}, \emph{off-policy}
algorithms can in principle learn from the data generated by any arbitrary source of behavior, making them ideal candidates
for batch learning.
However, these algorithms have been demonstrated to be unstable when learning from fixed datasets
with insufficient coverage due to overestimation bias in unfamiliar states~\cite{fujimoto2018off,qtopt}.

Batch RL has the potential to represent a significant step forward for robot learning, allowing robotics practitioners
to collect powerful calibration datasets of robot experience without requiring detailed task knowledge in advance, while enabling
completely offline training on arbitrary tasks that were not known at exploration time~\cite{bruce2017one}.
In this work, we would like to call attention to this relatively under-explored paradigm, and aim to take a step toward a solution by
evaluating the performance of various state-of-the-art exploration approaches 
for diverse task-agnostic experience collection, for offline learning of arbitrary tasks that were not known at dataset generation time.

\section{Related Work}

In this section we review the literature relating to off-policy learning on a fixed dataset, and the state-of-the-art in
exploration methods that might be used to produce a maximally diverse training dataset without knowing the ultimate tasks
of interest in advance.

\subsection{Off-Policy Learning}

Reinforcement learning algorithms can be broadly classified on a spectrum from \emph{on-policy} in which training data always comes from the
current version of the agent, to \emph{off-policy} in which the agent is able to learn from arbitrary experience; we are motivated
by learning completely offline, so we focus our attention on the latter.
Off-policy reinforcement learning has the potential to be particularly applicable in our situation, because it opens up the possibility of
learning from many sources of experience beyond that collected by the current policy~\cite{gu2017deep}.
Impressive results have been achieved in domains such as robotic grasping by making use of task-relevant datasets from previous
policies and even from entirely different experimental runs, in which diverse data collection was identified as an essential
requirement for offline learning~\cite{qtopt}.
These approaches can be susceptible to overestimation bias in unfamiliar states due to optimistic backup of estimated
future value; this bias can be mitigated to some degree with pessimism in the face of uncertainty by trusting the minimum of two
independent estimates~\cite{hasselt2010double,van2016deep,fujimoto2018addressing}.

In the case of far-off-policy learning, where the distribution mismatch grows very large between the training dataset and the
state-action visitation induced by the policy, this overestimation bias can lead to instability and complete learning failure
as estimation errors compound indefinitely without the possibility of correction by actively visiting those overestimated states online.
When the desired task is sufficiently similar to the task performed by the data-gathering policy, this issue can be mitigated by
Batch-Constrained Q-learning (BCQ), in which the policy is constrained to avoid distribution mismatch between the training and on-policy
data distributions. This approach is reminiscent of imitation learning, but with the benefit of being able to optimize arbitrary reward
functions at offline training time~\cite{fujimoto2018off}.

A method known as Goal Exploration Processes (GEP) has been proposed for the paradigm of explicit separation of task-agnostic
exploration and task-aware exploitation phases in reinforcement learning~\cite{colas2018gep}, in which randomized linear policies
are used to generate a bootstrap sample followed by a random perturbation procedure to encourage diversity of state visitation
as measured by hand-specified task-relevant features. The data gathered by GEP is then used to initialize the experience memory
of a state-of-the-art continuous RL agent~\cite{lillicrap2015continuous}, after which on-task training proceeded as usual.

In this paper, we consider a related but different paradigm in which the dataset is collected entirely in advance with no knowledge
of the ultimate tasks the agent will be trained on, and with no task-relevant features known ahead of time. Given this fixed dataset,
we then initialize a state-dependent reward function and train a task policy completely offline without any further interaction with
the environment. The need for extremely diverse data in advance in order to cover arbitrary future tasks puts extra pressure on the
exploration method, which forms the main focus of our evaluations in this paper.

\subsection{Exploration}

Learning generally requires exposure to diverse training data. In reinforcement learning, generating diverse training data
is typically achieved by an exploration mechanism internal to the agent in question, and exploration techniques have
been an active area of research in the field from early on~\cite{thrun1992efficient}. Exploration is usually considered in the
context of online learning, in which the agent must not only optimize its objective, but also take unexplored actions in order
to learn the consequences thereof. In this work, we are interested in exploration from a slightly different angle: how to generate
diverse datasets in the absence of any task feedback whatsoever.

Classical results show that the Q-learning algorithm is provably convergent in the tabular case, given complete exploration
of the problem~\cite{jaakkola1994convergence}. Although tabular guarantees no longer hold in the context of modern function
approximation, it is intuitive that effectively covering the space of the problem is important for convergent offline training.

The simplest and most common exploration techniques involve simply adding noise to the policy. The standard approach in discrete
Q-learning is known as $\epsilon$-greedy, in which a fraction $\epsilon$ of the time rather than acting optimally, the agent
chooses a random action~\cite{watkins1992q,mnih2015human}. In continuous control tasks similar noise-based exploration techniques
are often used, including directly adding \emph{iid} or correlated noise to the actions~\cite{lillicrap2015continuous,fujimoto2018addressing}.

Pure noise as a source of exploration behavior, while simple and requiring few assumptions, has difficulty reaching distant states:
the expected exploration time for a random policy to reach a given state grows exponentially with its distance, leading to the proposal
of \emph{deep exploration}~\cite{DeepExploration} in which an ensemble of policies are trained independently while sharing their experience,
resulting in consistent behavior policies that nonetheless result in diverse coverage of the problem space. GEP~\cite{colas2018gep},
described above, involves a similar technique in which a large number of randomized linear policies form the basis of the exploration behavior.

Another approach to exploration involves \emph{intrinsic motivation}~\cite{chentanez2005intrinsically}, in which the reward function of
the problem is augmented with an additive bonus that rewards the agent for visiting states in proportion to their novelty.
In count-based
exploration methods~\cite{bellemare2016unifying,tang2017exploration,ostrovski2017count}, the novelty of states is approximated directly
in inverse proportion to their visitation frequency.
An indirect way to measure familiarity is in terms of prediction error of a model
being trained in parallel with the RL agent, such as forward predictive models~\cite{pathak2017curiosity,pathak2019beyond} or the error in
predicting the state-dependent output of another arbitrary network~\cite{burda2018exploration}.
Diversity of states can be used directly
as an objective to optimize, by training a maximum-entropy RL agent to optimize its distinctiveness from other agents as measured by a
state-dependent classification network trained in parallel~\cite{eysenbach2018diversity}.
Intrinsic motivation has even been shown to
achieve impressive task performance in the complete absence of task reward, in situations in which pure exploration correlates with the
task objective~\cite{burda2018large}.

In self-driven learning more generally, model-based approaches can be trained purely self-supervised while providing analytic gradients to
train the policy directly~\cite{PILCO,DEEPPILCO,SVG}. Model ensembles can
be leveraged to provide an estimate of uncertainty in addition to diversity of experience~\cite{kurutach2018model}, and uncertainty-aware
methods like these could be used to backpropagate directly into a learning agent for either seeking or avoiding uncertainty~\cite{henaff2019model}
as the need arises.

In this work, we are primarily interested in task-agnostic exploration. We consider state-of-the-art exploration methods
Random Network Distillation (RND)~\cite{burda2018exploration},
Diversity Is All You Need (DIAYN)~\cite{eysenbach2018diversity},
and GEP~\cite{colas2018gep},
for the purpose of generating diverse datasets with no task knowledge, evaluated according to the performance of a separate off-policy
agent learning to optimize entirely new tasks unknown at exploration time.

\begin{figure*}[h!]
    \centering
    \begin{subfigure}[b]{0.2\textwidth}
        \includegraphics[width=\linewidth]{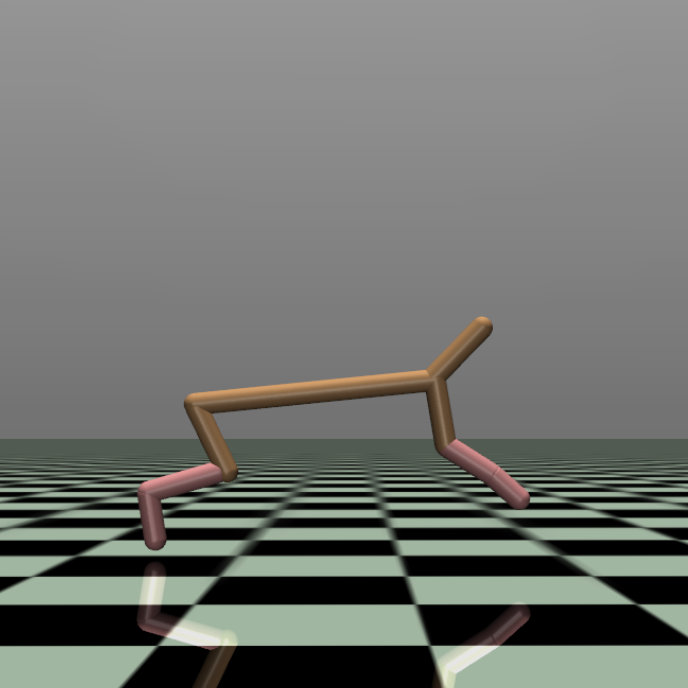}
    \end{subfigure}
    \begin{subfigure}[b]{0.2\textwidth}
        \includegraphics[width=\linewidth]{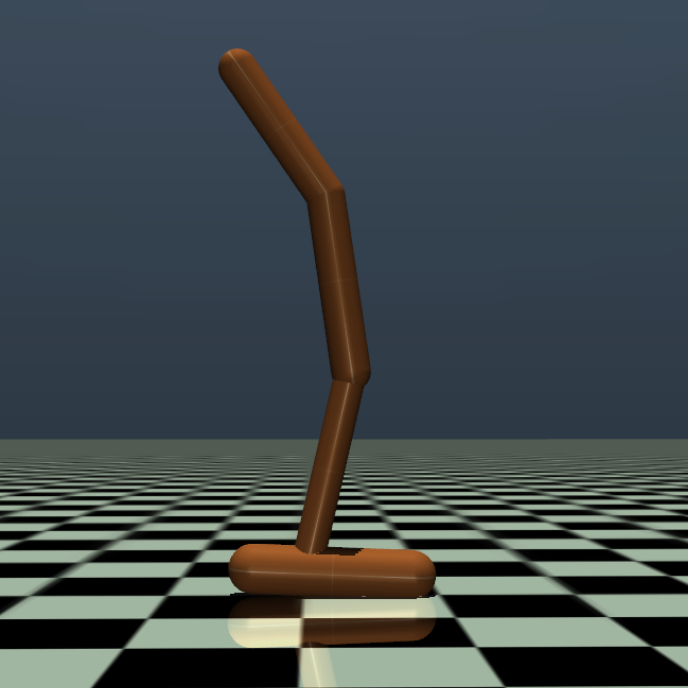}
    \end{subfigure}
    \begin{subfigure}[b]{0.2\textwidth}
        \includegraphics[width=\linewidth]{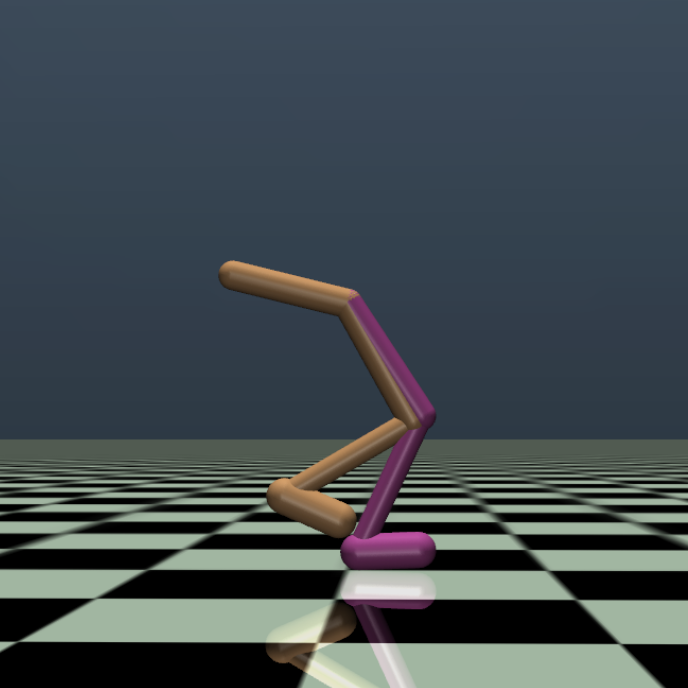}
    \end{subfigure}
    \begin{subfigure}[b]{0.2\textwidth}
        \includegraphics[width=0.9675\linewidth]{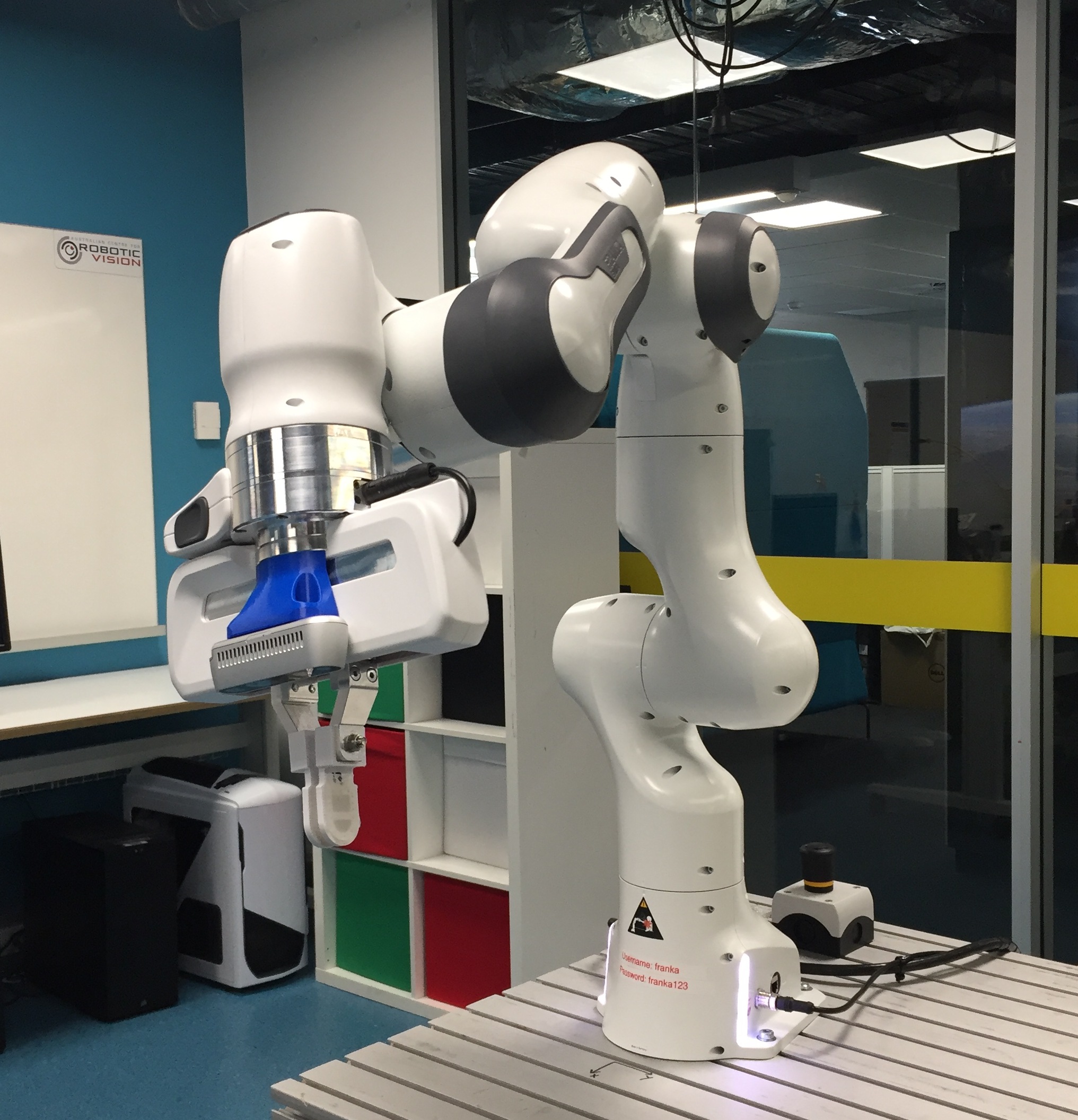}
    \end{subfigure}
    \caption{Experimental environments: HalfCheetah-v1, Hopper-v1, Walker2d-v1, and real FrankaEmika Panda arm with 7 degrees of freedom.}
    \label{fig:environments}
    \vspace{-0.3cm}
\end{figure*}

\section{Approach}
\label{sec:Approach}

In this work, we consider the problem of generating a static dataset of robot experience without task knowledge,
in order to prepare for learning to solve arbitrary tasks in the future completely offline. We decompose the problem
into two phases: \emph{exploration}, in which we execute a state-of-the-art exploration algorithm from the literature
for a fixed number of timesteps; and \emph{offline learning}, in which we train an off-policy RL algorithm to solve
arbitrary tasks that were not known to the exploration agent when the dataset was gathered.

\subsection{Exploration}
In this phase, we execute an exploration algorithm on the robotic platform for a fixed number of timesteps in order
to generate a dataset of diverse exploration data with no prior knowledge of the task. We describe three popular
exploration algorithms from the literature (RND, DIAYN, and GEP) as well as a simple baseline and a novel exploration
algorithm adapted from the literature.

\subsubsection*{\textbf{Random Network Distillation}}
In RND~\cite{burda2018exploration}, a randomly-initialized and fixed encoding function $f_\text{teacher}(x)\rightarrow{}\phi$ is used to encode
observations from the environment into fixed-length feature vectors. These feature vectors are used as the labels of
a supervised learning procedure to train another function $f_\text{student}(x)\rightarrow{}\Tilde{\phi}$. The reward
given to the exploration policy is the same objective that the supervised process is minimizing:

\begin{equation}
    R_\text{RND}(x_t) =  \| ~ \phi_{t} - \Tilde{\phi}_{t} ~ \|
\end{equation}

\subsubsection*{\textbf{Diversity Is All You Need}}
DIAYN~\cite{eysenbach2018diversity} is a reinforcement learning algorithm that trains an ensemble of diverse skills by rewarding each
policy for being distinct as measured by a learned classification algorithm $f_\text{d}(x) \rightarrow{} P(\text{skill} | x)$.
DIAYN trains an ensemble of maximum-entropy RL agents to maximize the following reward while acting as randomly
as possible:

\begin{equation}
    R_\text{DIAYN}(x_t) = \log P(\text{skill}_t | x_t)
\end{equation}

\subsubsection*{\textbf{Goal Exploration Processes}}
GEP~\cite{colas2018gep} attempts to gather diverse data by generating a set of randomized linear policies and executing them to
collect experience, which is stored in memory in the form of a task-dependent descriptor extracted from each trajectory.
Since we do not allow the exploration phase any knowledge of the ultimate tasks, we simply store the element-wise mean of the states along
the trajectory as its descriptor. Once a number of randomized policies have been executed ($N=50$ in our case, as in the original work),
random ``goal'' states are sampled from the state space, and the policy in memory associated with the nearest state to the goal is
perturbed with random noise and executed again, adding its new experience to the memory. We continue this procedure until our exploration
dataset is the desired size.

\subsubsection*{\textbf{Random Policies}}
To measure the importance of the goal-sampling step in GEP, we also evaluate a simple baseline in which only the randomized policy
step is applied. Rather than sampling goals and perturbing policies from memory, this baseline randomly initializes
a new policy every episode.

\subsubsection*{\textbf{Self-Supervised Exploration}}
In addition to the existing baselines, we evaluate a novel exploration approach obtained by turning the RL objective typically optimized
by prediction-error-based intrinsic motivation algorithms into a supervised objective through use of the forward model
$f_\text{forward}(x_t,a_t) \rightarrow{} \Tilde{x}_{t+1}$
that is often
trained as a byproduct of these approaches~\cite{pathak2017curiosity,pathak2019beyond}. The typical prediction-error-based reward
being maximized is of the form:

\vspace{-0.25cm}
\begin{equation}
    R_\text{intrinsic}(x_t,a_t,x_{t+1}) =  \| ~ x_{t+1} - \Tilde{x}_{t+1} ~ \|
\end{equation}
\vspace{-0.25cm}

RL algorithms usually account for the difficulty of predicting the long-term future by optimizing discounted rewards:

\vspace{-0.25cm}
\begin{equation}
    V_R = \sum\limits_{t=0}^{T} \gamma^t (1-D_t) R_{\text{intrinsic}}(x_t,a_t,x_{t+1})
\end{equation}
\vspace{-0.25cm}

where $\gamma \in [0,1)$ is the discount factor, and $D_t$ is an indicator variable describing whether or not the episode has ended by time $t$.
Because $R_t$ and $\Tilde{x}_{t+1}$ depend on the forward model that is a byproduct of these prediction-error based methods,
we can implement an exploration technique in which the gradient of the agent's policy is estimated by backpropagation directly
through our predictions of the future. We take inspiration from~\cite{pathak2019beyond}, but rather than assume the next state does not depend
on the action, we train a pair of forward models $f_{\{1,2\}}(x_t,a_t)\rightarrow \Tilde{x}_{t+1}$ and optimize for their divergence as a proxy for novelty.
Because the sum of future rewards also requires the episode termination variable, we also learn a termination
prediction model $f_\text{done}(x_t)\rightarrow \Tilde{D}_t$ online from data during exploration.
We train a Soft Actor-Critic (SAC)-style maximum-entropy RL agent~\cite{haarnoja2018soft} to directly maximize the following quantity, representing
the sum of intrinsic rewards plus entropy terms:

\vspace{-0.25cm}
\begin{equation}
    R_\text{SSE}(x_t,a_t) = \| ~ f_1(x_t,a_t) - f_2(x_t,a_t) ~ \|
\end{equation}

\vspace{-0.25cm}
\begin{equation}
    V_\text{SSE} = \sum\limits_{t=0}^{T}  \gamma^t (1-\Tilde{D}_t) ({R_{SSE}(x_t,a_t)} - \log(P(a_t)))
\end{equation}

When trained on simulated forward rollouts from the model and backpropagating through time, we obtain a model-augmented supervised version
of intrinsically-motivated SAC that we refer to as Self-Supervised Exploration (SSE).

\subsection{Offline Learning}
Given a fixed dataset of robot experience, we are interested in learning to solve arbitrary tasks completely
offline with no further interaction with the environment.
In this section, we describe the two off-policy algorithms that we evaluate for learning tasks offline on fixed data, both of
which are based on the deep deterministic policy gradient (DDPG)~\cite{lillicrap2015continuous}
algorithm for continuous Q-learning.

\subsubsection*{\textbf{Twin Delayed DDPG}}
TD3~\cite{fujimoto2018addressing} is an improvement to DDPG that aims to reduce the overestimation bias that is
common when training off-policy value functions. Two Q-networks are trained simultaneously, and the minimum
is chosen when evaluating the Q-value for the purpose of bootstrapping (as in ~\cite{hasselt2010double,van2016deep}),
which corresponds to pessimistic estimation in the face of uncertainty.
Furthermore, noise is added to the output of the policy during training to encourage smoothness of estimation
in small regions around observed experience. Finally, as in the original work, the policy is trained half as frequently as the
value networks.

\subsubsection*{\textbf{Batch-Constrained Deep Q-learning}}
BCQ~\cite{fujimoto2018off} achieves improved offline learning by training a state-conditional generative model
of the actions in the batch, which can then be used to sample actions that reflect the actions present in the dataset.
Keeping the policy action close to the buffer distribution reduces the extrapolation error that would otherwise accumulate
due to distribution mismatch and overestimation.

\section{Experiments}

In this work, we consider exploration methods to generate diverse datasets of robot experience for learning arbitrary tasks completely offline.
We first conduct a thorough evaluation of the approaches described in Section~\ref{sec:Approach} on three standard simulated continuous control
tasks as in~\cite{fujimoto2018off}. We then evaluate the best performing approaches to explore and then learn reaching tasks offline on a
physical FrankaEmika Panda robot arm with 7 degrees of freedom. All environments are shown in Figure~\ref{fig:environments}.

\subsection{Simulation Experiments}

We first evaluate on a standard benchmark suite of 3 OpenAI Gym MuJoCo environments: \verb|HalfCheetah-v1|, \verb|Hopper-v1|, and \verb|Walker2d-v1|~\cite{todorov2012mujoco,brockman2016openai}.
As described in Section~\ref{sec:Approach}, we partition the experiments into a task-agnostic exploration phase followed by a task-aware offline training phase.
In the exploration phase, each exploration method generates 1 million transitions of experience in the simulated domain in the form of $(x_t, a_t, x_{t+1})$, and this data is saved as a static dataset.

\begin{figure*}[h!]
    \centering
    \begin{subfigure}[b]{0.55\textwidth}
        \centering
        \includegraphics[width=\textwidth]{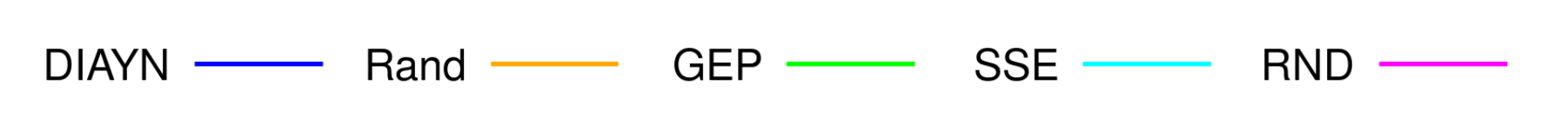}
    \end{subfigure}
    \begin{subfigure}[b]{0.49\textwidth}
        \includegraphics[width=\linewidth]{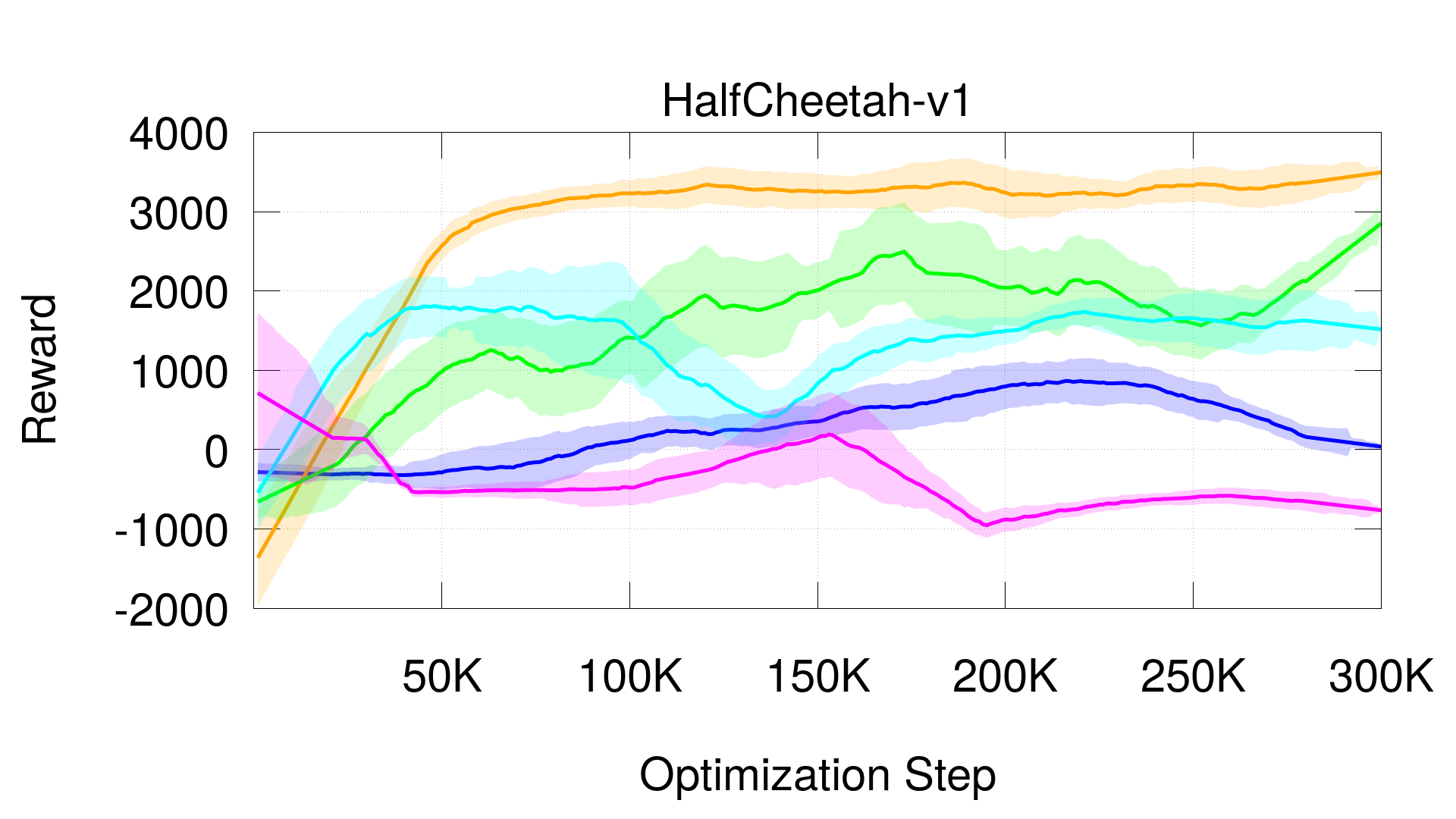}
    \end{subfigure}
    \begin{subfigure}[b]{0.49\textwidth}
        \includegraphics[width=\linewidth]{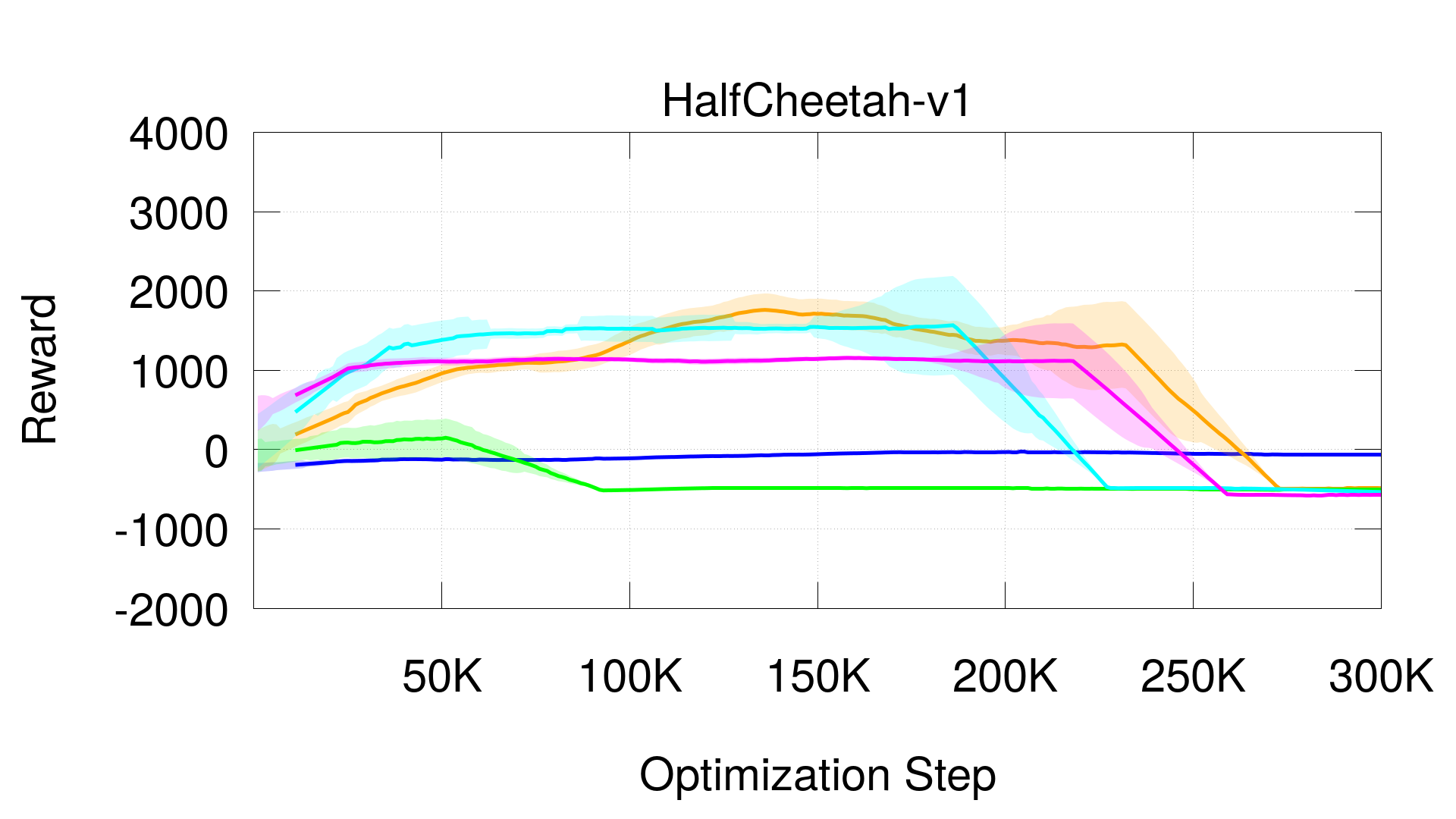}
    \end{subfigure}
    \begin{subfigure}[b]{0.49\textwidth}
        \includegraphics[width=\linewidth]{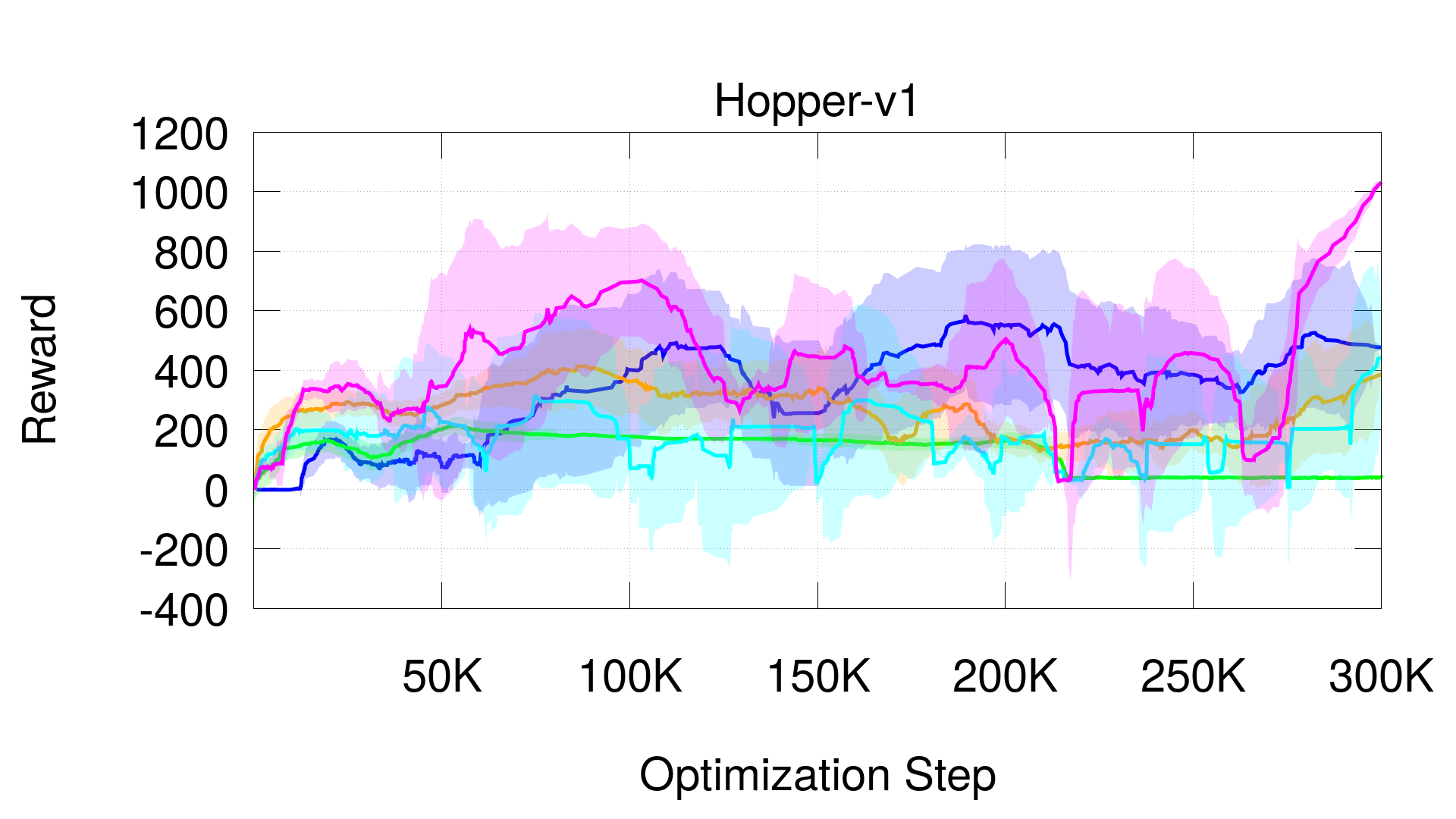}
    \end{subfigure}
    \begin{subfigure}[b]{0.49\textwidth}
        \includegraphics[width=\linewidth]{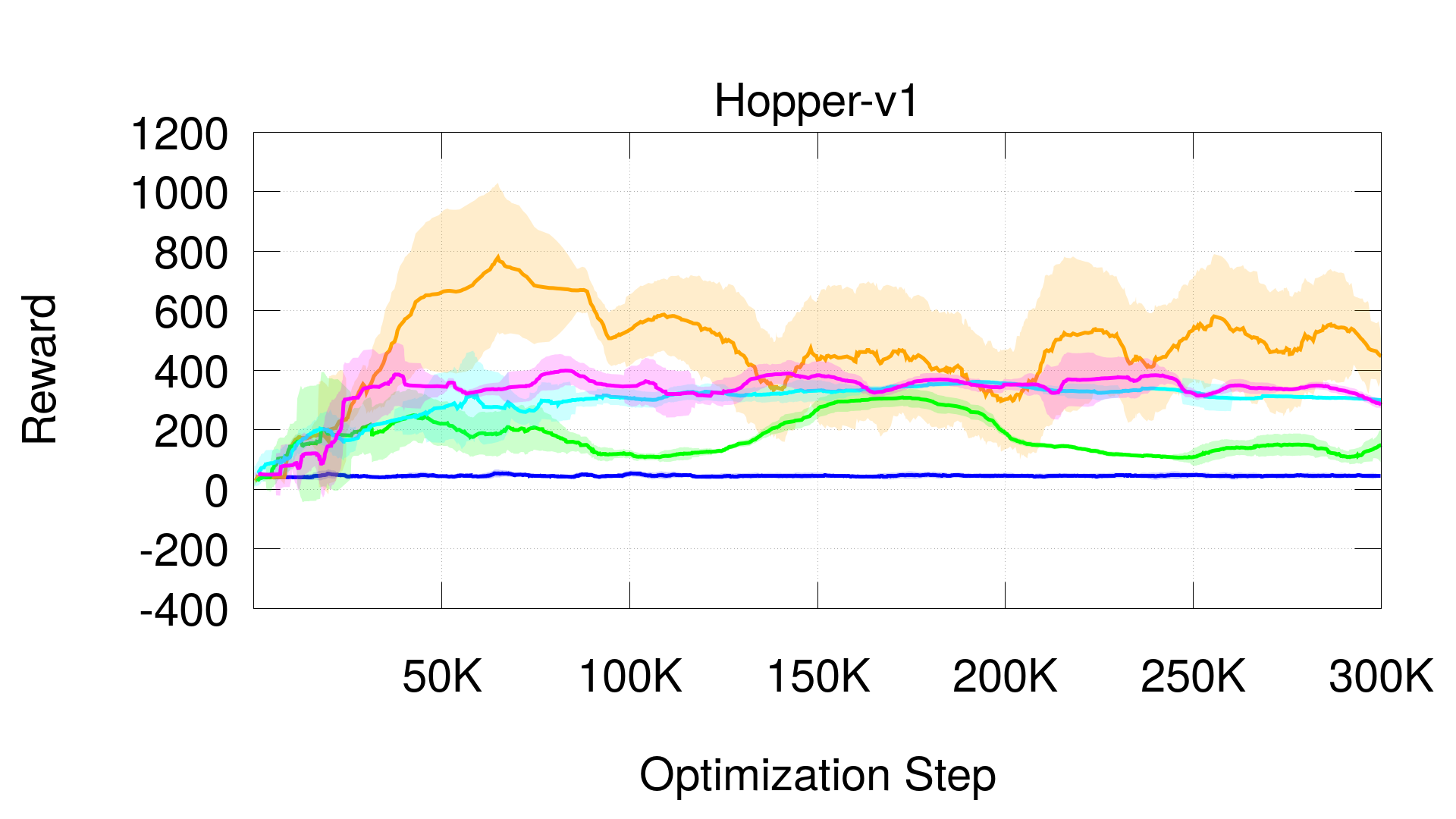}
    \end{subfigure}
    \begin{subfigure}[b]{0.49\textwidth}
        \includegraphics[width=\linewidth]{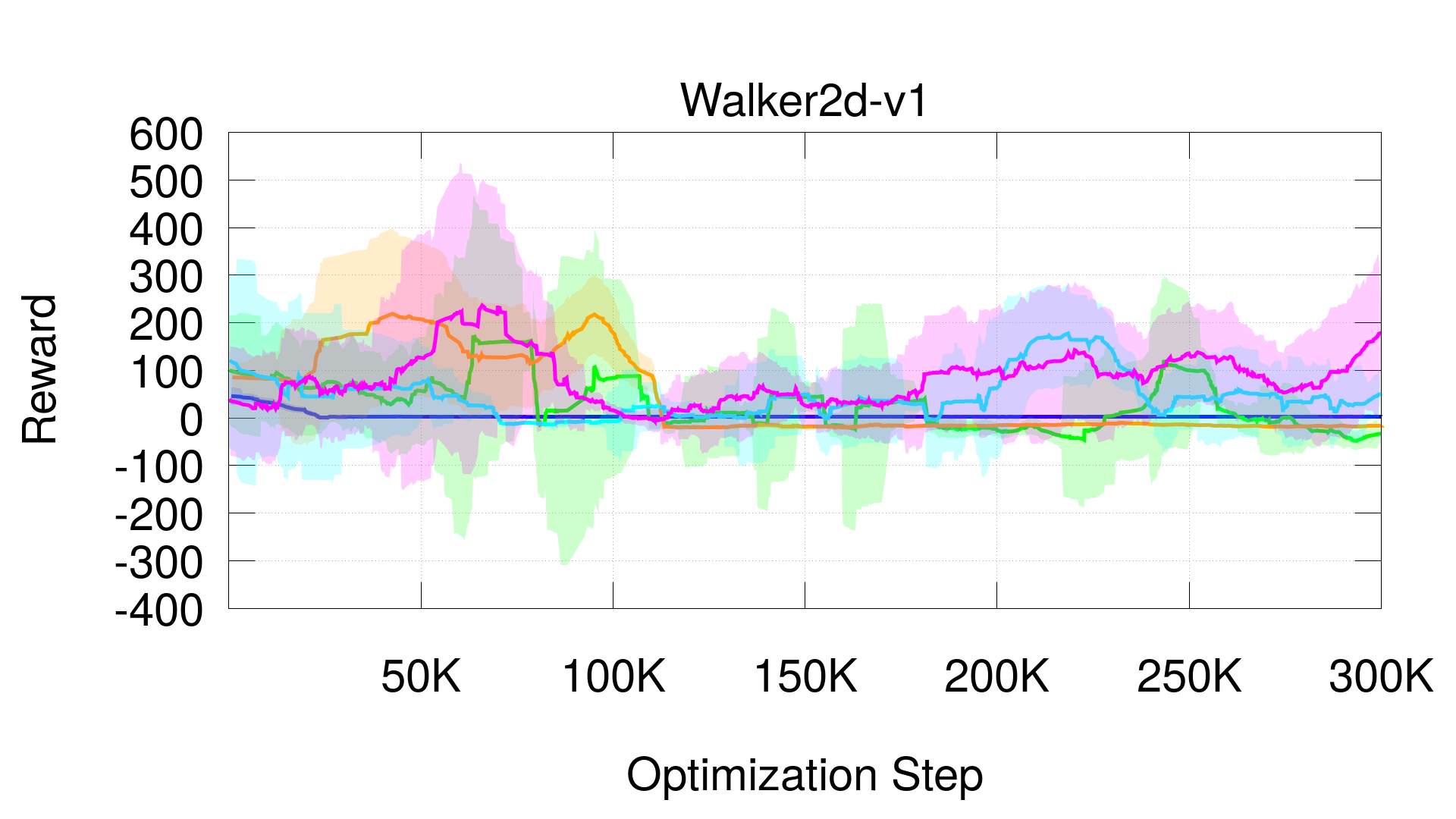}
        \caption{TD3}
    \end{subfigure}
    \begin{subfigure}[b]{0.49\textwidth}
        \includegraphics[width=\linewidth]{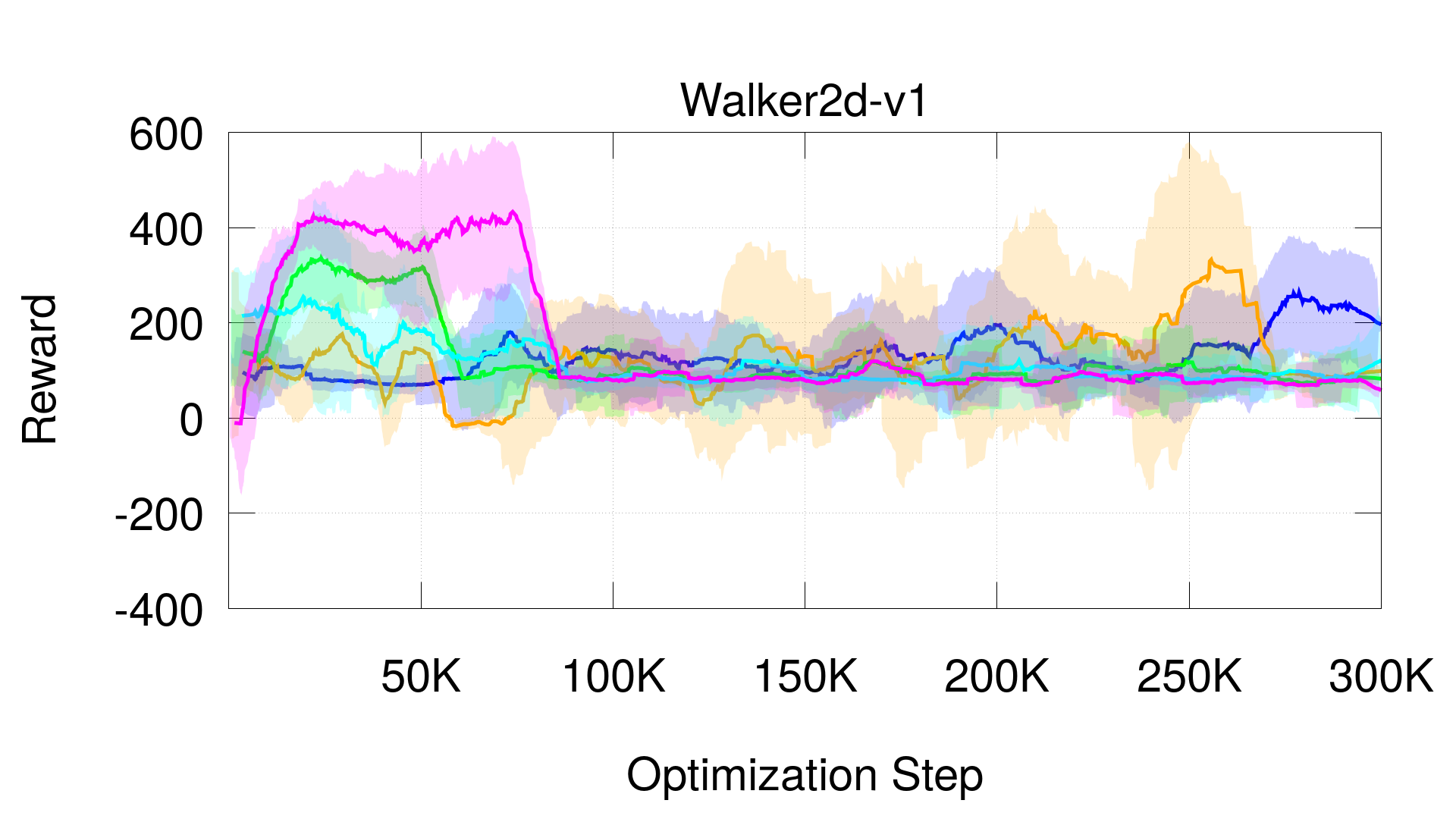}
        \caption{BCQ}
    \end{subfigure}
    \vspace{-0.3cm}
    \caption{Simulation Results}
    \label{fig:plots}
\vspace{-0.3cm}
\end{figure*}

In the offline training phase, we choose a task with a state-dependent reward function that was not known to the exploration agent, and train an off-policy RL algorithm to solve the task on the
data in the static dataset for 300,000 training steps. For task rewards, we use the standard locomotion tasks of maximizing forward velocity, as provided by the three environments.
We evaluate RND, DIAYN, GEP, random networks, and our proposal SSE as exploration methods, and TD3 and BCQ as off-policy learning algorithms.
In all cases we use the default hyperparameters from the original papers that proposed the algorithms, and code for our experiments is freely available at \url{https://github.com/qutrobotlearning/batchlearning}.
Results for the simulation experiments are shown in Figure~\ref{fig:plots}.

\subsection{Physical Robot}

In order to validate the approach on a real robot, we consider the FrankaEmika Panda robot arm platform with 7 degrees of freedom. Agents observed joint angles in radians, and joint velocities in radians per second,
resulting in observation vectors of length 14. Actions were sent at 20Hz in the form of joint velocities in radians per second, clipped in the range $[-0.5, 0.5]$, and episodes were reset after 1000 timesteps or
when the robot violated physical safety limits such as self-collision.
We collected data on the physical robot in a manner similar to the simulated domains, but we limited the dataset size to 200K transitions due to the additional time cost of physical experiments.
Policies were trained using TD3 and BCQ offline to solve reaching tasks starting from a deterministic ``home'' configuration to one of four different goals, specified differently for each task.
Results for the real robot experiments are shown in Table~\ref{tab:real_world_results}, comparing the distance
to the target point for each exploration method and offline learning algorithm.

\begin{table}[]
\begin{tabular}{@{}crrrr@{}}
\toprule
\textbf{Learning Method} & \multicolumn{4}{c}{\textbf{Exploration Method}} \\ \cmidrule(l){2-5} 
                         & Rand  & DIAYN          & RND   & SSD            \\ \midrule
TD3                      & 0.46  & 0.51           & 0.49  & \textbf{0.30}  \\
BCQ                      & 0.47  & \textbf{0.30}  & 0.52  & 0.44           \\ \bottomrule
\end{tabular}
\caption{Distances of closest tooltip positions in meters for each evaluated method on the real robot.}
\label{tab:real_world_results}
\vspace{-0.3cm}
\end{table}

\section{Discussion}

Somewhat counter-intuitively, the state-of-the-art RL exploration methods we evaluated did not perform particularly
well in our experiments, as shown in the HalfCheetah-v1 results in Figure~\ref{fig:plots}.
Particularly surprising is the result that randomly generating a new linear policy every
episode seems to outperform many of the other baselines by a wide margin. This suggests that despite achieving impressive 
results during online training, current methods of exploration are not well suited to the pure exploration paradigm, as described in this paper. Furthermore, BCQ did not perform as well as expected, but this is reasonable as it was not designed to
learn from purely task-agnostic data. Also of interest is that the best performing algorithms on the real robot did correspond to the
best performance on the simulation tasks.
Note however that we did not engage in heavy
parameter tuning of the offline learning algorithms we used, and Rainbow-style improvements~\cite{hessel2018rainbow}
to off-policy algorithms may provide improvements to the result regardless of the exploration method used.

During the exploration phase on the real robot we observed that the randomly generated linear policies seemed to generate vastly diverse actions that in totality covered a larger portion of the state space compared to the other exploration methods. This may imply that while systematically covering the state space might be useful for exploration given an unlimited dataset, with the restrictions of a limited static dataset it is vital to explore regions in the state space that are far apart. The greater diversity in the dataset may increase generalization capabilities of the agent to nearby previously-unexplored states while reducing the chances of visiting states completely different from those in the dataset.

\section{Conclusion}

In this work, we consider the paradigm of task-agnostic exploration for generating datasets that are diverse enough
to train policies to solve arbitrary tasks with no further interaction with the environment. This is an important
goal for robotics, potentially enabling a single diverse dataset to train robots with a lifetime worth
of skills on demand.

Our experiments showed interesting and unexpected results for the state-of-the-art exploration methods and off-policy algorithms in this setting.
Since exploration has been shown to be an important component of RL performance, we were expecting the established exploration
algorithms to generate diverse enough data to train tasks offline, but in domains such as Hopper-v1 and Walker-2d,
purely self-directed exploration without a task seems to be very challenging. 
We believe that this justifies further research in this paradigm given the potential benefits to robotics
from single-dataset offline training.

We make our algorithms, experiments, and hyperparameters freely available on \url{https://github.com/qutrobotlearning/batchlearning}.







\bibliographystyle{named}
\bibliography{bibliography.bib}

\end{document}